\documentclass[letterpaper, 10 pt, conference]{ieeeconf}  % Comment this line out if you need a4paper

\usepackage{url}
\usepackage{graphicx}
\usepackage{booktabs}
\usepackage{romannum}
\usepackage{amsfonts}
\usepackage{multirow}
\usepackage[table,xcdraw]{xcolor}
\usepackage{array}
\usepackage{float}
\usepackage[export]{adjustbox} 
\usepackage{algorithm}
\usepackage{balance}
\usepackage{cite}
\usepackage{algpseudocode} 
\usepackage{longtable}
\usepackage{booktabs}

\usepackage{graphics}
\usepackage{upgreek}
\usepackage{amssymb}
\usepackage{refstyle}

\usepackage{amsmath}
\usepackage{subfigure}
\usepackage{todonotes}
\usepackage{mathtools}

\usepackage{hyperref}

\usepackage{cleveref}

\newcommand\copyrighttext{%
  \footnotesize \textcopyright 2023 IEEE. Personal use of this material is permitted.
  Permission from IEEE must be obtained for all other uses, in any current or future
  media, including reprinting/republishing this material for advertising or promotional
  purposes, creating new collective works, for resale or redistribution to servers or
  lists, or reuse of any copyrighted component of this work in other works.
  DOI: \href{https://ieeexplore.ieee.org/document/10354958}{No. 10.1109/ROBIO58561.2023.10354958}} %  \href{https://ieeexplore.ieee.org/document/10354958}{No. 10.1109/ROBIO58561.2023.10354958}
\newcommand\copyrightnotice{%
\begin{tikzpicture}[remember picture,overlay]
\node[anchor=south,yshift=10pt] at (current page.south) {\fbox{\parbox{\dimexpr\textwidth-\fboxsep-\fboxrule\relax}{\copyrighttext}}};
\end{tikzpicture}%
}

\IEEEoverridecommandlockouts                              % This command is only needed if
\title{\LARGE \bf
A Closed-Loop Multi-perspective Visual Servoing Approach\\ with Reinforcement Learning 
}

\author{Lei Zhang$^{1,2}$\dag, Jiacheng Pei$^{2,3}$\dag, Kaixin Bai$^{1,2}$, Zhaopeng Chen$^{2,1}$*, Jianwei Zhang$^{1}$% <-this % stops a space
\thanks{\dag The first two authors contribute equally to this paper.}% <-this % stops a space
\thanks{*Corresponding author.}
\thanks{{$^{1}$TAMS (Technical Aspects of Multimodal Systems), Department of
Informatics, University of Hamburg}, {$^{2}$Agile Robots AG}, {$^{3}$RWTH Aachen University}}
\thanks{** This research has received funding from the German Research Foundation (DFG) and the National Science Foundation of China (NSFC) in project Crossmodal Learning, DFG TRR-169/NSFC 61621136008, partially supported by European projects H2020 STEP2DYNA (691154) and ULTRACEPT (778602).}
}
\DeclarePairedDelimiterX{\norm}[1]{\lVert}{\rVert}{#1}

\begin{document}

\maketitle
\copyrightnotice

\thispagestyle{empty}
\pagestyle{empty}

%%%%%%%%%%%%%%%%%%%%%%%%%%%%%%%%%%%%%%%%%%%%%%%%%%%%%%%%%%%%%%%%%%%%%%%%%%%%%%%%
\begin{abstract}
Traditional visual servoing methods suffer from serving between scenes from multiple perspectives, which humans can complete with visual signals alone. In this paper, we investigated how multi-perspective visual servoing could be solved under robot-specific constraints, including self-collision, singularity problems. We presented a novel learning-based multi-perspective visual servoing framework, which iteratively estimates robot actions from latent space representations of visual states using reinforcement learning. Furthermore, our approaches were trained and validated in a Gazebo simulation environment with connection to OpenAI/Gym. Through simulation experiments, we showed that our method can successfully learn an optimal control policy given initial images from different perspectives, and it outperformed the Direct Visual Servoing algorithm with mean success rate of $97.0\%$.
\end{abstract}

%%%%%%%%%%%%%%%%%%%%%%%%%%%%%%%%%%%%%%%%%%%%%%%%%%%%%%%%%%%%%%%%%%%%%%%%%%%%%%%%
\section{INTRODUCTION}

Humans can guide their behavior using semantic information from visual images captured at different angles. Similarly, visual servoing enables robots to adjust their motion based on visual feedback. Visual servoing is influenced by several factors, including feature matching and robot trajectory planning. However, traditional visual servoing methods often assume that the initial observation pose is similar to the target's observation pose, limiting their applicability in multi perspective visual servoing scenarios.

Multi-perspective visual servoing in complex industrial scenarios poses greater challenges due to difficulties in acquiring accurate object models and the inherent complexity of scenes. Occlusions of objects also sharpen the difficulties of pose estimation, making 6D pose-based visual servoing inadequate for such situations. In contrast, traditional visual servoing methods are expert in achieving high-speed, low-error robot guidance within local regions based on image features or specific features. However, precise servoing to the target position from arbitrary poses remains challenging. Hand-crafted features are often used in traditional visual servoing methods, which leads to traditional visual servoing being applicable only to small convergence domains. 
\begin{figure}[htbp]
	\begin{center}
		\includegraphics[width=6.7cm]{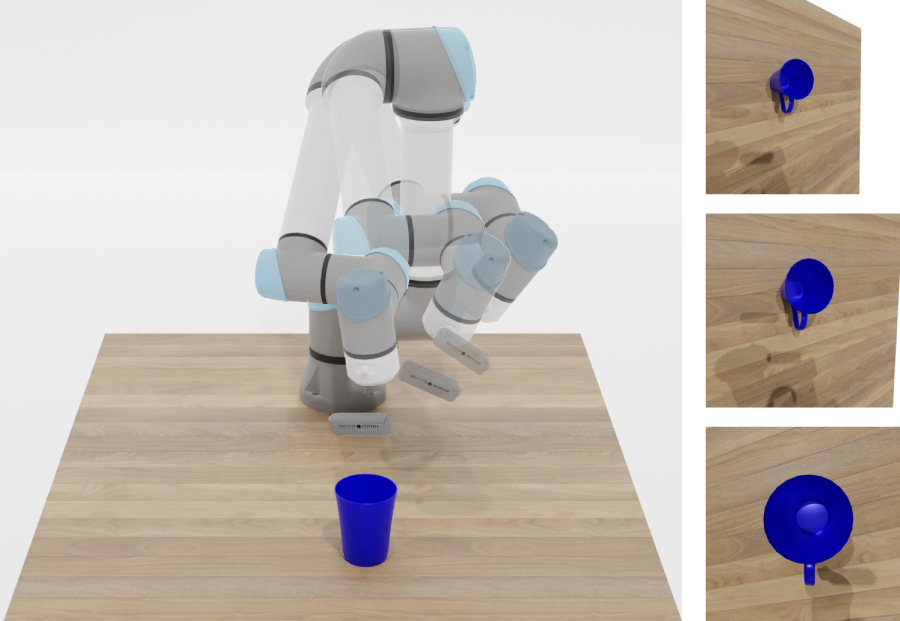}
		\caption{The sythetic robot agent is controlled with proposed closed-loop visual servoing approach to target pose with desired visual state.}
		\label{fig.task-diagram}
	\end{center}
\end{figure}
Significant disparities between the image features of the initial state and the target state result in increased challenges in feature matching of visual servoing. Simultaneously, the constraints of the robot need to be taken into account during visual servoing. Ultimately, the difficulties in multi-view visual servoing persist due to perceptual limitations and the constraints imposed by the robot.

Recently learning-based visual servoing has also been studied. Deep learning-based approaches have shown enhanced performance in dealing with complicated scenarios~\cite{saxena2017exploring}, occluded environments, and varying lighting conditions~\cite{bateux2018training,yu2019siamese}. Reinforcement learning were also utilized in improving generalization of visual servoing~\cite{sampedro2018image,shi2020adaptive,singh2019model,lampe2013acquiring,shi2016decoupled,jin2021policy}. However, most of the current learning-based visual servoing approaches only consider 2D perspectives and do not address the problem of robotic operations under robotic constraints. Training robust robot control strategies under robot-specific constraints are also the core problem of deep learning-based robot manipulation methods. 
In this work, we consider the problem of multi-perspective visual servoing using reinforcement learning (RL) methods, as shown in Fig.~\ref{fig.task-diagram}. Our method utilizes autoencoder network to extract latent space representations of current and desired camera sensor data. We propose a closed-loop robot control policy network to estimate robot action from the latent space representations. 
Our core contributions are:
\begin{itemize}
\item A closed-loop multi-perspective visual servoing framework utilizing RL to servo robot agent from latent space representations of visual states.
\item Improve training efficiency of RL-based robotic policy using learning from demonstration method and Hindsight Experience Replay (HER)~\cite{andrychowicz2017hindsight}.
\item A potential-based reward function with consideration of the task- and robot-specific constraints.
\end{itemize}

\section{RELATED WORK AND BACKGROUND}\label{RELATED WORK AND BACKGROUND}

\subsection{Visual Servoing}
Image-based visual servoing (IBVS) \cite{weiss1987dynamic,feddema1989vision} uses the extracted 2D features as states to achieve robot control, whereas pose-based visual servoing (PBVS) \cite{wilson1996relative, thuilot2002position} estimates the poses of camera in the Cartesian space and guides the robot by minimizing pose error. 
IBVS suffers from convergence and stability problems due to ill-conditioned image Jacobian matrix and singularities~\cite{chaumette1998potential} and the problem of finding optimal path in the Cartesian space~\cite{corke2001new}.
 PBVS is limited by the image quality and calibration errors. 
The traditional methods require tracking a set of handcrafted features, such as points, lines or patterns. 
Direct or photometric visual servoing (DVS) treats the luminance of image pixels as the visual features and computes the interaction matrix~\cite{deguchi2000direct,Kallem2007KernelbasedVS,collewet2008visual,dame2009entropy,collewet2011photometric}. However, DVS suffers from small convergence domain. It may be failed when features are not visible or in challenging lighting condition~\cite{bateux2018training}.

Recent advances in machine learning open up new opportunities to improve the flexibility, robustness,
and accuracy of existing visual servoing methods. 
Supervised learning-based VS, such as KOVIS~\cite{puang2020kovis} and Siame-se(3)~\cite{felton2021siame}, leverages a network to extract features for observed and target images, such as key points or latent space features, and subsequently utilizes the intermediate results to train a policy network to predict motion of robotic arm. Siame-se(3)~\cite{felton2021siame} utilized PBVS in simulation for collecting dataset. However, these approaches to opportunistic deep learning mostly focus on the perceptual part and finding the optimal path for multi-view visual servoing is still rarely investigated.

\subsection{Reinforcement Learning-based VS and Continuous Control}

RL offers advantages over deep learning for visual servoing, including adaptability to uncertainty, support for long-term decision-making. However, predicting continuous action spaces using RL is expected to become more challenging due to the exponential growth in the size of continuous action spaces compared to discrete spaces. Lillicrap et al.~\cite{lillicrap2015continuous} proposed off-policy algorithm named Deep Deterministic Policy Gradients (DDPG) with models of actor and critic. To address the overestimation problem of DDPG, Fujimoto et al.~\cite{pmlr-v80-fujimoto18a} proposed an algorithm called Twin Delayed Deep Deterministic Policy Gradient (TD3). 

To learning optimal VS control policy with less dependency on prior domain knowledge and tackle unexpected disturbances, RL is utilized to train visual servoing control law~\cite{sampedro2018image}. Sampedro et al.~\cite{sampedro2018image} used DDPG algorithm to build an IBVS controller for multirotor aerial robots. Shi et al.~\cite{shi2020adaptive} combined Q-learning with fuzzy state coding to adjust the image Jacobian matrix in IBVS efficiently and adaptively, aiming to stabilize and improve the VS performance for wheeled mobile robots. Singh et al.~\cite{singh2019model} showed that prioritized experience replay buffer could improve the convergence time of RL-based IBVS. Lampe et al.~\cite{lampe2013acquiring} proposed reinforcement learning-based visual servoing for reaching and grasping.

\section{PROBLEM STATEMENT AND METHOD}\label{PROBLEM STATEMENT AND METHOD OVERVIEW}
\subsection{Problem statement}

We formulate the reinforcement learning-based visual servoing as a Markov decision process. The robot agent executes a continuous action $\boldsymbol{a}_t$ estimated by robot control policy, denoted as $\pi_{\rm vs}$, according to any given state $\boldsymbol{s}_t$ at time $t$. 
After action execution, reward is calculated based on updated new state $\boldsymbol{s}_{t+1}$. The main objective of RL-based visual servoing is to train an optimal robot control policy $\pi^{*}_{\rm vs}$ that maximizes the expected rewards $R$ in the future. 

In our work, we investigate a robot visual servoing off-policy to estimate robot joint velocities $\boldsymbol{\dot{q}}_{t+1}$ based on state $\boldsymbol{s}_t$. The policy aims to iteratively achieve the desired pose by maximizing the object function $J$ with respect to state $S$ and keeping robot free of singularities and self-collisions, as demonstrated in Eq.~\ref{equ:RL_problem_statement}.

\begin{equation}
\label{equ:RL_problem_statement}
\begin{array}{c}
\boldsymbol{a}=\pi_{\rm vs}\left(\boldsymbol{s} ; \boldsymbol{w}^p\right) \\
J\left(\boldsymbol{w}^p\right)=\mathbb{E}_S\left[\widehat{q}\left(\boldsymbol{s}, \boldsymbol{a} ; \boldsymbol{w}^v\right)\right]
\end{array}
\end{equation}
where $\boldsymbol{w}^p$ and $\boldsymbol{w}^v$ denote the weights of policy network (actor) $\pi_{\rm 
vs}$ and value network (critic) $\widehat{q}$. 

\begin{figure*}[htbp]
	\begin{center}
		\includegraphics[width=15cm]{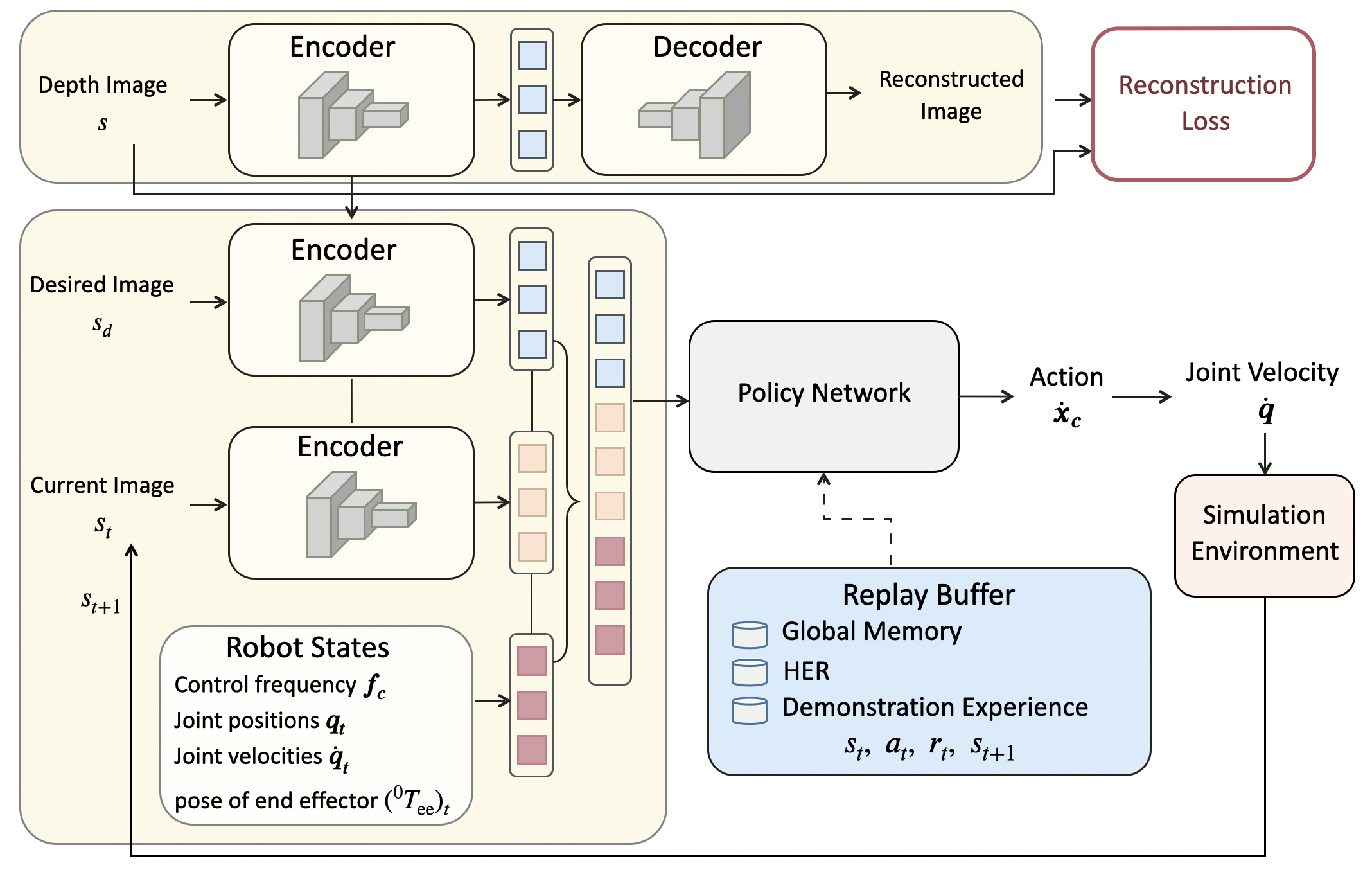}
		\caption{Architectures of Autoencoder and Closed-Loop Multi-Perspective Visual Servoing Network. Top Part: The latent space representations are extracted with encoder of autoencoder. Bottom Part: The robot policy estimates joint velocities based on latent space visual representations of current and desired image frames and robot states, including joint position, joint velocities, end-effector pose, and control frequency. The action of joint velocities is executed in simulation environment and the policy network is trained using reinforcement learning.}
		\label{fig.network-architecture}
	\end{center}
\end{figure*}

The main architecture of RL-based visual servoing is shown in Fig.~\ref{fig.network-architecture}. Firstly, the latent space representations are learned from visual state, as introduced in Sec.~\ref{subsection:method-learn-laten-space-representation}. Then, proposed closed-loop multi-perspective visual servoring network estimated robot actions based on latent space representations and robot states, as detailed in Sec.~\ref{subsection:visual-servoing-policy-network}. 

\subsection{Learning Latent Space Representations from Autoencoder}
\label{subsection:method-learn-laten-space-representation}
To acquire generalized features and minimal information of images for visual servoing, an autoencoder network is utilized to extract latent space representations from depth image. The autoencoder is trained with reconstruction loss function based on self-supervison:
\begin{equation}
    \mathcal{L}(\boldsymbol{w}_1, \boldsymbol{w}_2; \boldsymbol{y})=\mathbb{E}\left[\left(\boldsymbol{y}-\underbrace{\mathrm{P}_{\boldsymbol{w}_1}\left(Q_{\boldsymbol{w}_2}(\boldsymbol{y})\right.}_{\boldsymbol{y}^{\prime}})\right)^2\right]
\end{equation}
where $\boldsymbol{y}$ represents the network input, network output $\boldsymbol{y}^{\prime}$ is calculated based on encoder $\mathrm{P}_{\boldsymbol{w}_1}$ and decoder $Q_{\boldsymbol{w}_2}$ with weights $\boldsymbol{w}_1$ and $\boldsymbol{w}_2$.

\subsection{Closed-Loop Multi-Perspective Visual Servoing Robotic Policy Network}
\label{subsection:visual-servoing-policy-network}
\subsubsection{Action Space and Observation Space}
The action space of proposed reinforcement learning-based visual servoing is the velocity of camera frame with minimal and maximal limits, as described as:
\begin{equation}
   \begin{aligned} \mathcal{A} &=\left\{{ }^{c} \dot{\boldsymbol{x}}_{c} \mid{ }^{c} \dot{\boldsymbol{x}}_{c} \in\left[{ }^{c} \dot{\boldsymbol{x}}_{c, \min },{ }^{c} \dot{\boldsymbol{x}}_{c, \max }\right]\right\} \\{ }^{c} \dot{\boldsymbol{x}}_{c} &=\left(v_{c, x}, v_{c, y}, v_{c, z}, \omega_{c, x}, \omega_{c, y}, \omega_{c, z}\right) \end{aligned} 
\end{equation}
where $\mathcal{A}$ denotes the action space of RL algorithm. $\dot{\boldsymbol{x}}_{c}$ represents camera frame velocity, $v_{c, x}, v_{c, y}, v_{c, z}$ and $\omega_{c, x}, \omega_{c, y}, \omega_{c, z}$ denote linear and angular speeds in $x$, $y$, $z$ direction.

The observation space $\mathcal{S}$ consists of the visual states expressed in latent space $\boldsymbol{S}_{t}, \boldsymbol{S}_{\rm des}$ and the robot states $\boldsymbol{S}_{\rm robot}$:
\begin{equation}
    \mathcal{S}=\left\{\boldsymbol{S}_{t}, \boldsymbol{S}_{\rm des}, f_{\rm c}, \boldsymbol{q}, \dot{\boldsymbol{q}},{ }^{0} \boldsymbol{x}_{\rm e e}\right\}
\end{equation}
where $f_{\rm c}$ denotes control frequency, $\boldsymbol{q}$ and $\dot{\boldsymbol{q}}$ represent joint positions and joint velocities. ${ }^{0} \boldsymbol{x}_{\rm e e}$ denotes the Cartesian pose of end-effector coordinate frame.
\subsubsection{Architecture}
Firstly, the real-time captured image $\boldsymbol{I}_{t}$ at time $t$ and desired image $\boldsymbol{I}_{\rm des}$ are fed into encoders with shared weights to obtain the latent space features $\boldsymbol{S}_t$ and $\boldsymbol{S}_{\rm des}$. Secondly, the robot policy network takes the $\boldsymbol{S}_{\rm des}$, $\boldsymbol{S}_{t}$ together with $\boldsymbol{S}_{\rm robot}$ as inputs. The robot states encompass control frequency $f_{\rm c}$, joint positions $\boldsymbol{q}$ and velocities $\boldsymbol{\dot q}$, as well as the pose
${}^0\boldsymbol{T}_{\rm ee}$ of the end-effector derived from the robot's forward kinematics. The network then produces Cartesian space velocity $\boldsymbol{\dot{x}}_{\rm c}$ of camera coordinate frame and based on this velocity, we calculate joint velocities $\boldsymbol{\dot{q}}_{t}$ as follows:
\begin{equation}
\begin{array}{cc}
         {v}_{\rm ee} = {}^{\rm ee}\boldsymbol{V}_{\rm c} \boldsymbol{\dot{x}}_{\rm c}  \\
      \boldsymbol{\dot{q}} = ({}^{\rm e}\boldsymbol{J}_{\rm e})^{-1} {v}_{\rm ee}
\end{array}
\end{equation}
where ${}^{\rm ee}\boldsymbol{V}_{\rm c}$ and ${}^{\rm e}\boldsymbol{J}_{\rm e}$ denote twist transformation matrix and Jacobian matrix, ${v}_{\rm ee}$ introduces velocity of end effector.

After robotic movement based on joint velocities $\boldsymbol{\dot{q}}_{t}$ and control frequency $f_{\rm c}$, the camera image $\boldsymbol{I}_{t+1}$ and robot states $f_{\rm c}, \boldsymbol{q}_{t+1}, \boldsymbol{\dot q}_{t+1}, ({}^0\boldsymbol{T}_{\rm ee})_{t+1}$ are updated and serve as the feedbacks in the closed control visual servoing loop.

\subsubsection{Twin-Delayed Deep Deterministic-based Policy Gradient Agents}
\begin{figure*}[htbp]
	\begin{center}
		\includegraphics[width=15cm]{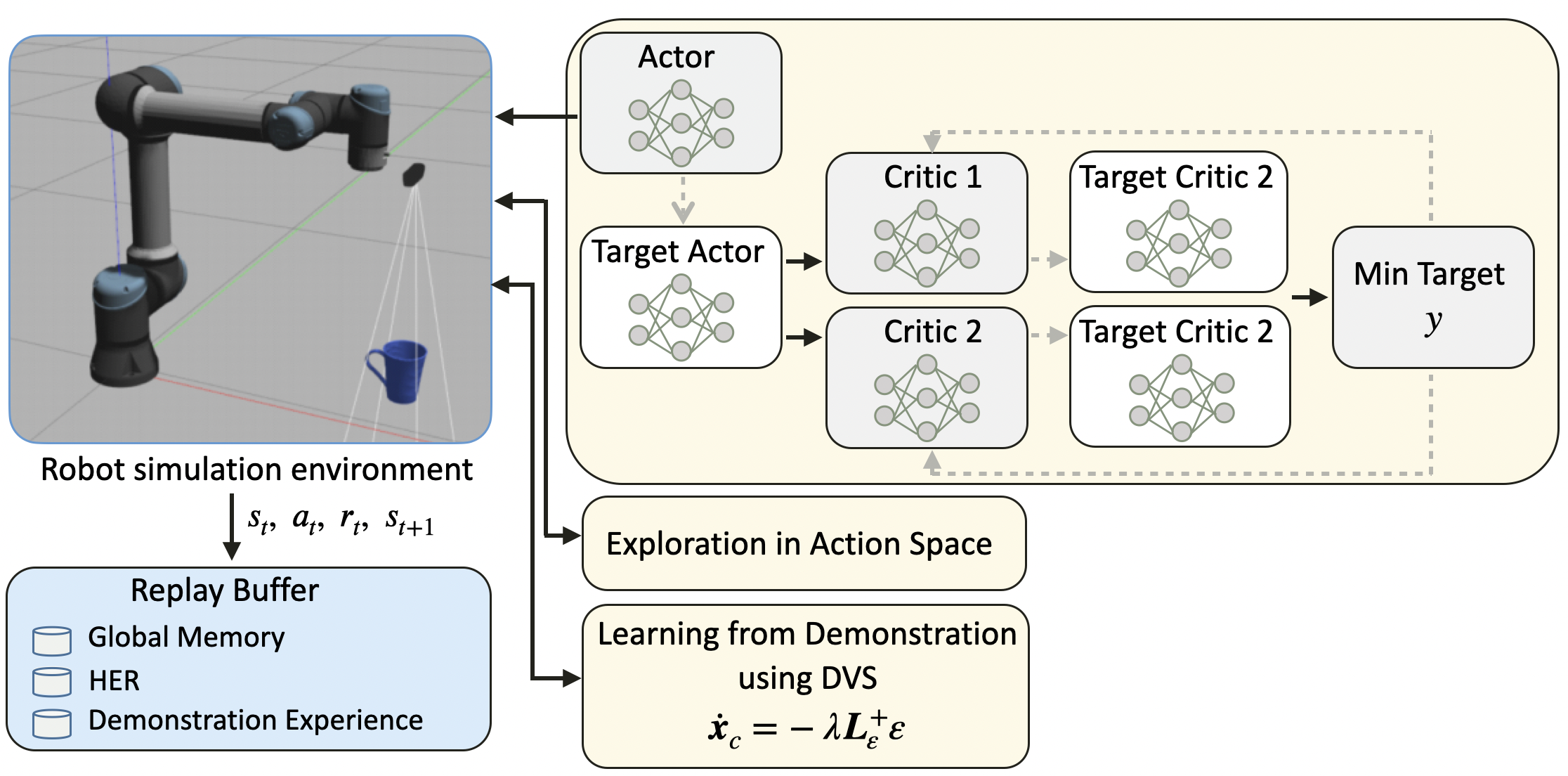}
		\caption{Architecture of proposed visual servoing policy network. Exploration from action space is utilized to capture more trajectory data. Traditional DVS method is used to generate imperfect success demonstration experience and HER is applied in generating experiences from achieved goals.}
		\label{fig.policy-network-architecture}
	\end{center}
\end{figure*}

We design our robot policy based on the architecture of the actor-critic network TD3. It incorporates a total of six networks, comprising two critic neural networks, one actor neural network, and three target neural networks, as detailed in Fig.~\ref{fig.policy-network-architecture}.

TD3 is proposed based on clipped double Q-learning structure, delayed updates of policy networks, and target networks to solve the overestimation problem. 
The policy network (actor) estimates the velocities of camera frame and the robot action $\boldsymbol{\widehat{a}}_{t+1}$ is executed in our simulation environment. 
\begin{equation}
    \boldsymbol{\widehat{a}}_{t+1}=\pi\left(\boldsymbol{s}_{t+1} ; \boldsymbol{w}^p\right)+\epsilon
\end{equation}
where $\boldsymbol{w}^p$ represents weights of the policy network and $\epsilon$ denotes the noise from a clipped normal distribution.

Based on the output $\widehat{q}\left(\boldsymbol{s}, \boldsymbol{a} ; \boldsymbol{w}^v\right)$ of value network (critic) $Q_\pi\left(\boldsymbol{s}, \boldsymbol{a} ; \boldsymbol{w}^v\right)$ , temporal difference target (TD-target) $\widehat{y}_t$ is calculated as follows:
\begin{equation}
    \widehat{y}_t=r_t+\eta \widehat{q}\left(\boldsymbol{s}_{t+1}, \boldsymbol{\widehat{a}}_{t+1}; \boldsymbol{w}^v\right)
\end{equation}
where $r_{t}$ denotes reward and $\boldsymbol{w}^v$ represents weights of value network. The value networks are updated based on error of the TD-target $\hat{q}\left(\boldsymbol{s}_t, \boldsymbol{a}_t ; \boldsymbol{w}^v\right)-\hat{y}_t$. During updating parameters, the minimal TD-target is selected from two target critics. Taking the minimal TD-target $y$ can not only reduce the overestimation but also mitigate the bias propagation. The update of target policy networks is less frequent than the update of critic networks to avoid using highly variant value estimates to update the policy. The delayed update improves the performance of policy networks and stabilizes the training process. In each episode, the transition $\left(\boldsymbol{s}_t, \boldsymbol{a}_t, r_t, \boldsymbol{s}_{t+1}\right)$ will be stored in the replay buffer as a global memory part and sampled as a train set.

\subsubsection{Reward function}

We propose reward function $r_t$ considering task- and robot-specific constraints for multi-perspective visual servoing, as formulated in Eq.~\ref{equ:reward}. The reward function considers the translation error $e_{\rm trans}$, rotational error $e_{\rm rot}$, and image difference error $e_{\rm img}$, as well as error related to training step $e_{\rm step}$ and terminal reward $r_{\rm terminal}$ with weights $\phi_{1},\phi_{2},\phi_{3},\phi_{4}$.
\begin{equation}
\label{equ:reward}
    \begin{aligned} r_{t}=& \phi_{1}\left(e_{\rm {trans }, t}-e_{\rm {trans }, t+1}\right)+\phi_{2}\left(e_{\rm {rot }, t}-e_{\rm {rot }, t+1}\right)+\\ & \phi_{3}\left(e_{\rm {img }, t}-e_{\rm {img }, t+1}\right)-\phi_{4} e_{\rm {step}}+r_{\rm {terminal}}, \end{aligned}
\end{equation}

If the robot faces singularity, collision, joint limitation problems, the training will be terminated and transition is stored. The terminal reward is formulated as follows.
\begin{equation}
\begin{array}{lll}
        r_{\rm terminal}&=&\left\{
        \begin{array}{ll}
        100-\lVert\boldsymbol{a}_{t}\rVert, & e_{\rm {trans}}<\varphi_{\rm trans} \wedge \\
        &
        e_{\rm {rot }}<\varphi_{\rm rot} \wedge \neg \rm { failure } ; \\ -100, &\text {if failure},

\end{array} \right. \\

\text{failure} &= &true,~\text{if} 
\left \{
   \begin{array}{l}
       e_{\rm {trans }}>\varphi_{\rm trans} \vee \\ e_{\rm {rot }}>\varphi_{\rm rot} \vee \\ \operatorname{det}\left({ }^{0} \boldsymbol{J}_{e e}\right) \leq \varphi_{\rm {Jacobian }} \vee \\ \rm {joint limits reached } \vee \\ \text {collision detected } \vee \\ \text {object out of FOV } \vee \\ \text {max. step reached }
       \end{array}
\right.

\end{array}
\label{equ:terminal_reward}
\end{equation}
where $e_{\rm {trans}, t}$ and $e_{\rm {rot}, t}$ denote translation error and rotational error between camera pose ${ }^{0} \boldsymbol{T}_{\rm c}$ and ${ }^{0} \boldsymbol{T}_{\rm c^{*}}$ at time step $t$. $e_{\rm{img}, t}$ represents image difference between $I_{t}$ and $I_{\rm des}$. 
$e_{\rm{step}}$ introduces constant step error with value $1$. The binary terminal reward of $\pm 100$ is set such that the average episode returns of $+100$ and $-100$ could represent the success and failure respectively. The $\varphi_{\rm {trans}}$,~$\varphi_{\rm {rot}}$,~$\varphi_{\rm {Jacobian}}$ mean thresholds of translation error, rotation error, and Jacobian matrix.

\subsubsection{Training Strategy.}
\label{subsection:method-train-strategy}
To investigate the efficiency of training RL-based policy for robotic visual servoing with sparse action space, four variants of our proposed method were trained according to following training strategies:

\textbf{Hindsight Experience Replay. }
To improve the sample efficiency of RL algorithm with sparse reward and sparse data, HER is proposed by learning from failed experiences. Using HER method, achieved states are sampled from failed trajectory as reachable goals and the transitions are stored into replay buffer. Specifically, the policy could learn to achieve an arbitrary given goal from HER experiences.

\textbf{Learning from Demonstration. }
Due to sparse distribution of success action of visual servoing, it's challenging to learn proposed policy using pure TD3. We propose the method to move the robot agent nearby the desired state and utilize traditional DVS method to collect imperfect demonstration experiences, according to:
\begin{equation}
    \dot{\boldsymbol{x}}_c=-\lambda \boldsymbol{L}_{\varepsilon}^{+} \varepsilon .
\end{equation}
where $\dot{\boldsymbol{x}}_c$ denotes camera velocity, $\boldsymbol{L}_{\varepsilon}^{+}$ represents Moore-Penrose pseudo-inverse of interaction matrix. $\varepsilon$ denotes visual error and the weight $\lambda$ denotes a positive scalar value.

\textbf{Additional Exploration. } We add additional exploration phase before training TD3-based network, where the random actions are sampled before train and replay buffer collects the corresponding transitions with exploration experiences.

Based on above training strategies, four variants are introduced:

\textbf{Pure TD3.} Select pure TD3 as baseline method.

\begin{table}[]
    \centering
    \begin{tabular}{|c|c|c|c|}
    
    \hline & Setting 1 & Setting 2 & Setting 3 \\
    \hline Range of object movement & $\pm 5 \mathrm{~cm}$ & $\pm 10 \mathrm{~cm}$ & $\pm 10 \mathrm{~cm}$ \\
    \hline Random initial pose & No & No & Yes \\
    \hline
    \end{tabular}
    \caption{Experimental settings based on different range of object movement and initial pose.}
    \label{tab_experiment_configuration_scenarios}
\end{table}
\begin{figure}[htbp]
	\begin{center}
		\includegraphics[width=8cm]{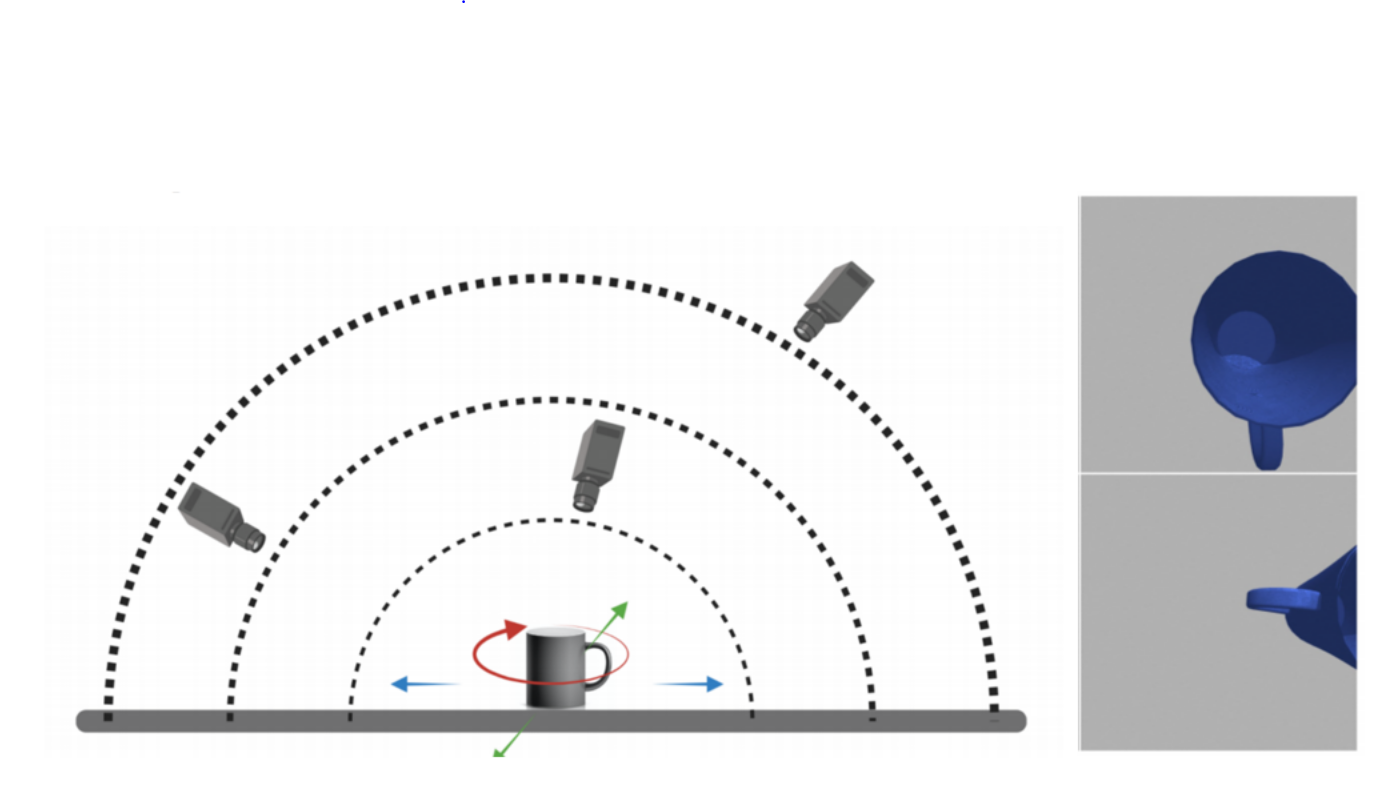}
		\caption{Scene generation for training the autoencoder and rendered images.}
		\label{fig.data-generation-encoder}
	\end{center}
\end{figure}

\textbf{TD3 + Exploration.} The robot policy network is based on the pure TD3 network and additional exploration phase is added before parameter updating of policy and value networks.

\textbf{TD3 + HER + exploration.} At the beginning of each episode in both the exploration and the training phase, we generated additional experiences in replay buffer based on HER method.

\textbf{TD3 + HER + exploration + Learning from Demonstration.} The imperfect demonstration experiences are generated with a certain probability and stored into replay buffer.

\section{EXPERIMENTS}\label{EXPERIMENTS}
We trained proposed closed-loop visual servoing network and executed evaluation experiments in simulation environment. The experiments aimed to $1)$ investigate reinforcement learning method in multi-perspective visual servoing with sparse data distribution and rewards, $2)$ compare the traditional visual servoing method and proposed method. %

Sec.~\ref{subsection:exp_setup_data_collection} describes the experimental setup and pipeline of data collection. Sec.~\ref{subsection:exp_simulation_exp_and_result} details the comparison experiments of our proposed methods and baseline methods. 

\subsection{Experimental Setup and Data Collection}
\label{subsection:exp_setup_data_collection}
The experimental setup was shown in Fig.~\ref{fig.policy-network-architecture}, where UR5e robot agent observed the blue cup on the table with Intel RealSense D435 camera. Before starting visual servoing, the observed object was placed on the table in a range and randomly rotated along z-axis. As well, we designed three experimental settings based on the range of the object and whether the robot's initial pose was randomized or not before visual servoing, as summarized in Tab.~\ref{tab_experiment_configuration_scenarios}.

\begin{figure}[htbp]
	\begin{center}
		\includegraphics[width=8.5cm]{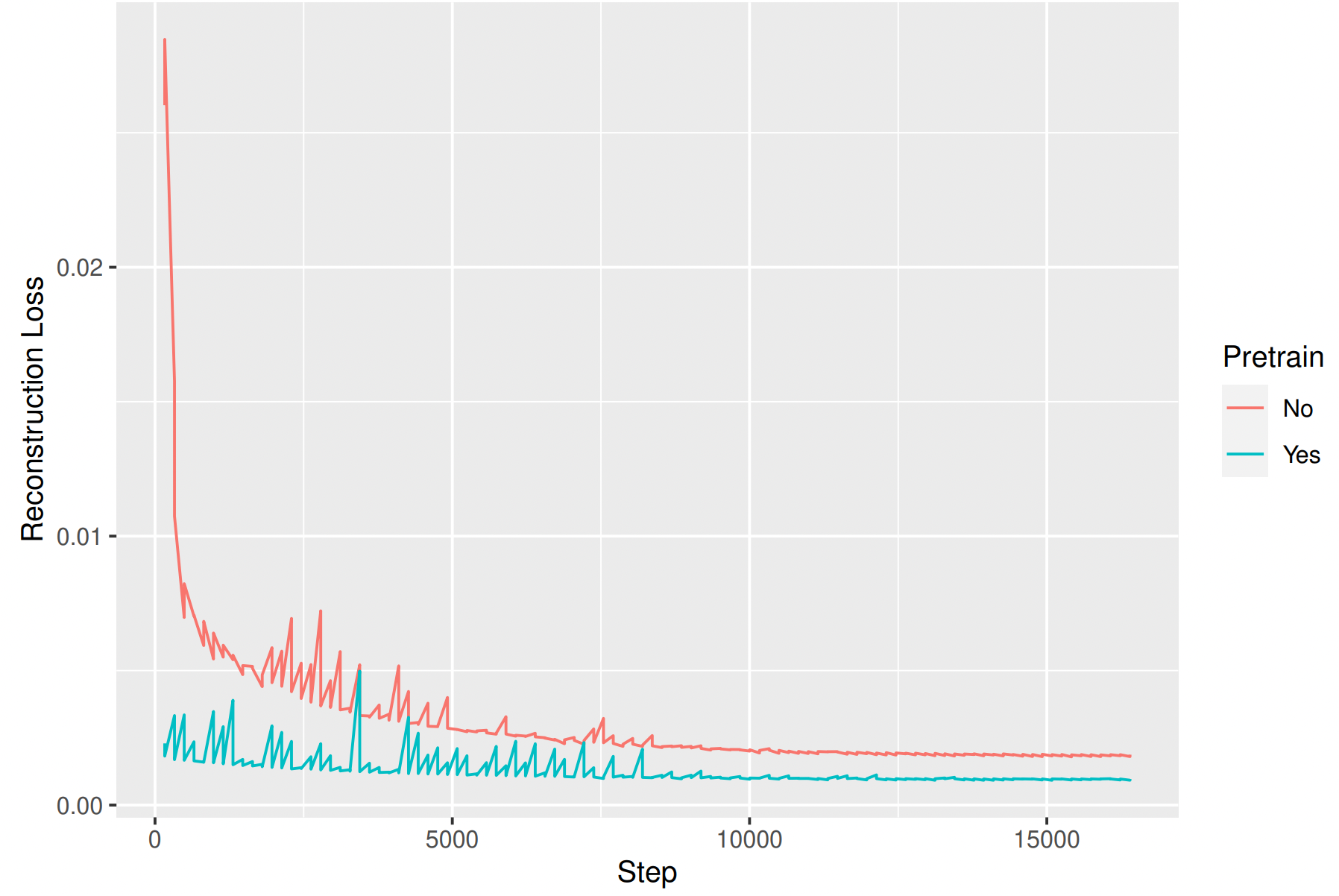}
		\caption{Training curves of autoencoder with and without pre-trained model.}
		\label{fig.autoencoder-train-curve}
	\end{center}
\end{figure}

To learn visual servoing from latent space representation, we first trained an autoencoder. To achieve this, we generated a dataset where depth and color images are rendered from a wide range of different relative poses between the object and camera. 
As shown in Fig.~\ref{fig.data-generation-encoder}, we generated the random poses of the camera on an upper spherical surface with a random radius. Subsequently, we moved the object pose randomly along the x- and y-axis and rotate it around the z-axis.  

During data generation, 100 random poses of the camera were generated and 100 random object poses were synthesized for each camera pose. The radius was selected from $50\rm{mm}$ to $850\rm{mm}$. Finally, one dataset with 10,000 samples was generated to train the autoencoder to learn the latent space representation of the object from different perspectives. We trained network with pre-training model from CelebA dataset~\cite{liu2015deep} and the training curves with and without pre-trained model were shown in Fig.~\ref{fig.autoencoder-train-curve} to demonstrate successful convergence of autoencoder.

After training the autoencoder, the synthetic robot agent was controlled based on the proposed visual servoing method. During the episodes of the training procedure, the simulation environment, value and policy networks, and HER buffer were firstly initialized. The goal state was randomly generated where the observed cup was in the observation area of the synthetic camera. In the training process with maximal step, policy network sampled action $\boldsymbol{a}_t$ and state $\boldsymbol{s}_{t+1}$ were collected after execuation of robot agent with velocity control. %

\begin{figure}[htbp]
	\begin{center}
    \includegraphics[width=8cm]{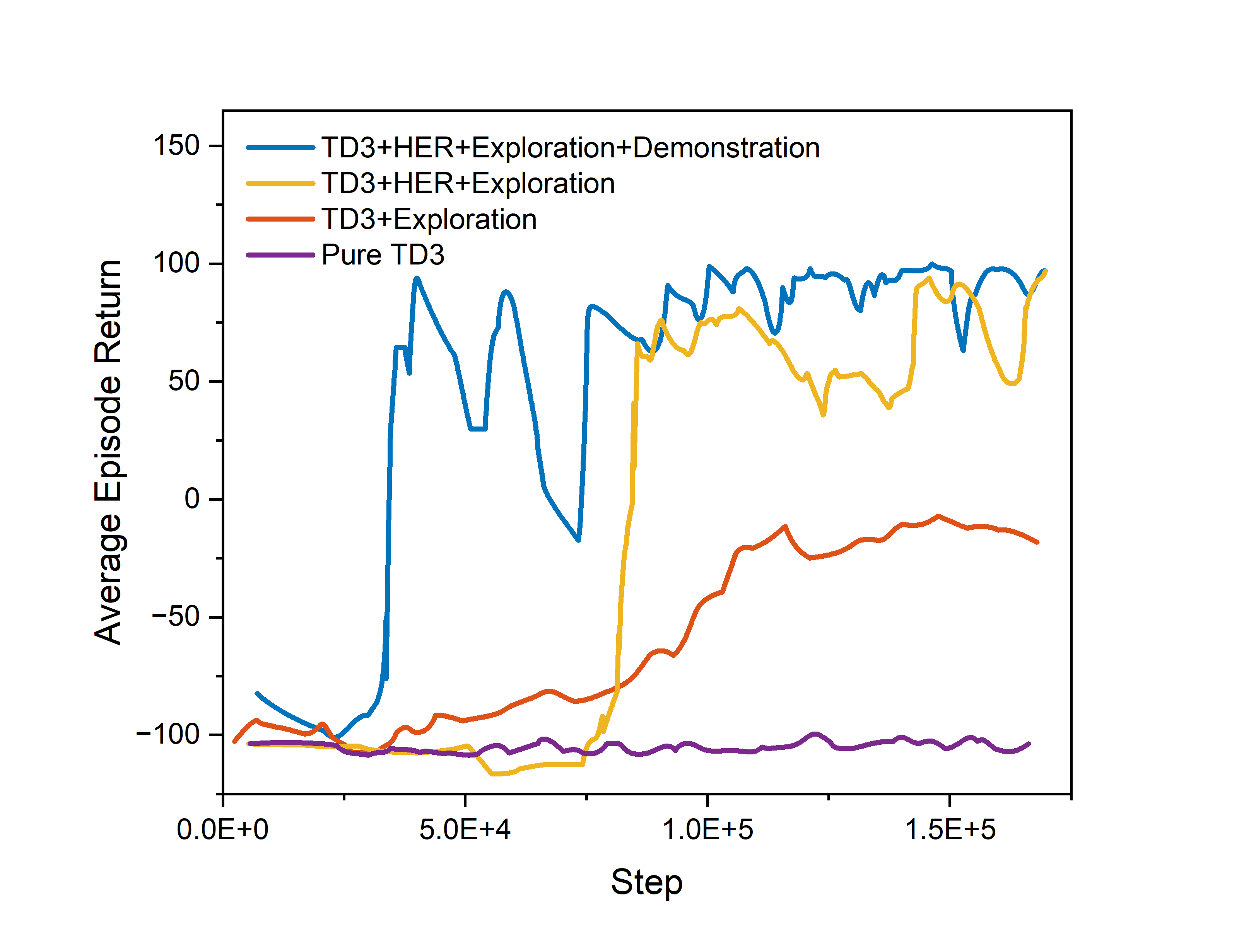}
		\caption{Training curves of four variants in first experimental configuration.}
		\label{fig.comparison_experiment_TD3_HER_exploration_demonstration}
	\end{center}
\end{figure}

\subsection{Simulation Experiments of Visual Servoing}
\label{subsection:exp_simulation_exp_and_result}
We executed comparison experiments of the four variants of proposed method, comparison experiments of our proposed method and traditional baseline method DVS.

In comparison experiments of four variants of proposed methods, the first experiment setting was applied during training. The training curves were shown in Fig.~\ref{fig.comparison_experiment_TD3_HER_exploration_demonstration}. The variant TD3+HER+Exploration+Demonstration performed best performance. The pure TD3-based network could not converge for multi-perspective visual servoing from latent space representations. The exploration method slightly improved the performance of the pure TD3-based variant. HER-based variants can converge successfully. This suggests that HER is beneficial in mitigating the problem of sparse data and reward function. The variant using learning from demonstration could converge faster.

In comparison experiments of our method and traditional baseline method, we conducted 100 comparison experiments under each experimental setting. We chose DVS as the traditional baseline method due to the lack of distinctive visual features on the observed object. 

Evaluation metrics were considered, including success rate and error distribution in translation and rotation. The visual servoing was counted as success when the error was smaller than a threshold. During training, we set the threshold as $2\rm{mm}$. Secondly, the translation and rotation errors were summarized based on a series of experiments in simulation.

\begin{figure}[htbp]
	\begin{center}
		\includegraphics[width=8cm]{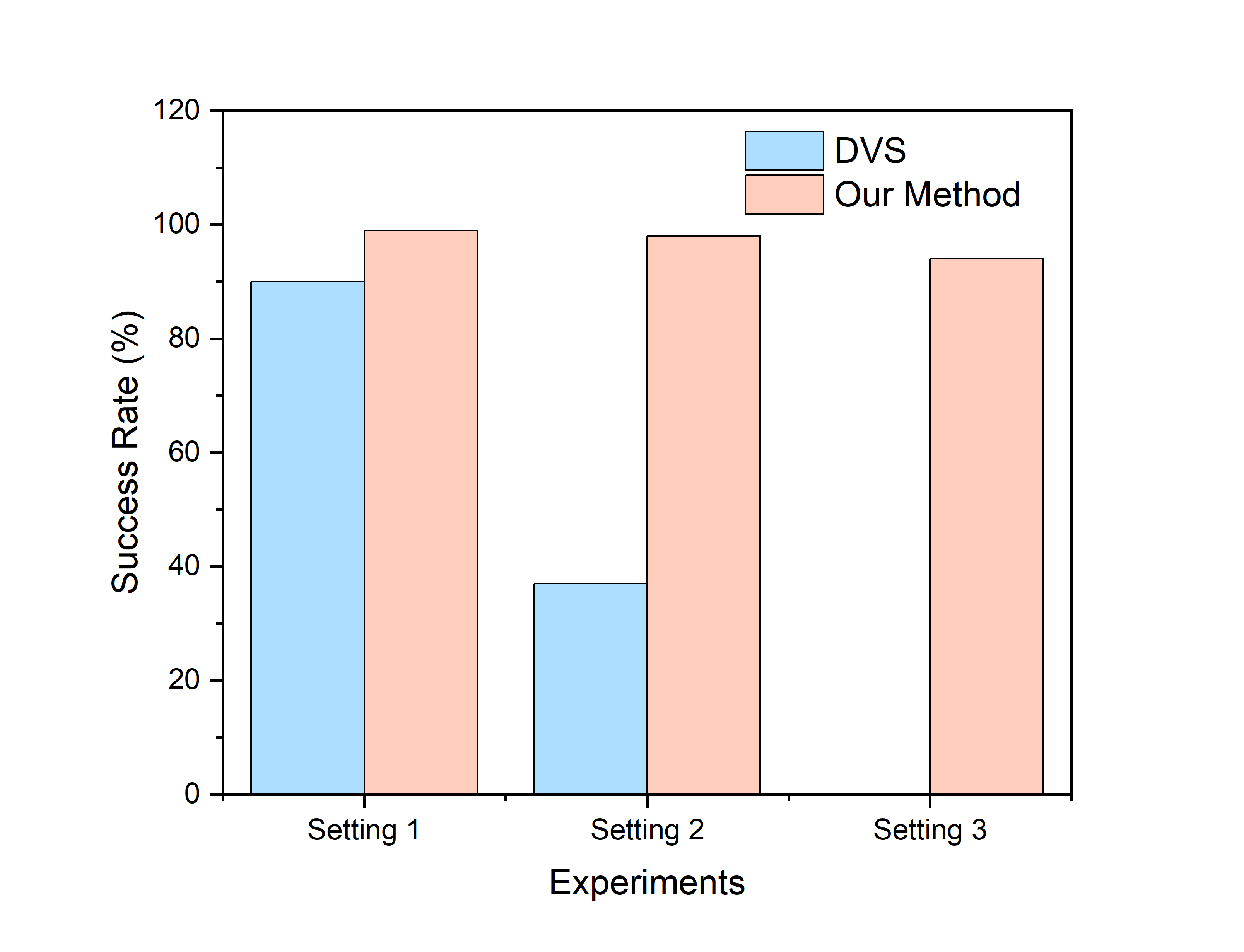}
		\caption{Success rates of DVS and our method in different experimental settings. The vacant bar within the bar chart symbolizes a success rate of 0\%.}
		\label{fig.comparison_our_and_dvs_success_rate}
	\end{center}
\end{figure}

Our method significantly outperformed DVS in terms of success rate. The mean success rate of the proposed method achieved above $97.0\%$ in different experimental settings, as shown in Fig.~\ref{fig.comparison_our_and_dvs_success_rate}. Meanwhile, the traditional visual servoing method DVS performed a poor success rate in complicated experimental settings. The mean success rate of experimental settings achieved $42.3\%$.

\begin{figure}[htbp]
	\begin{center}
		\includegraphics[width=8cm]{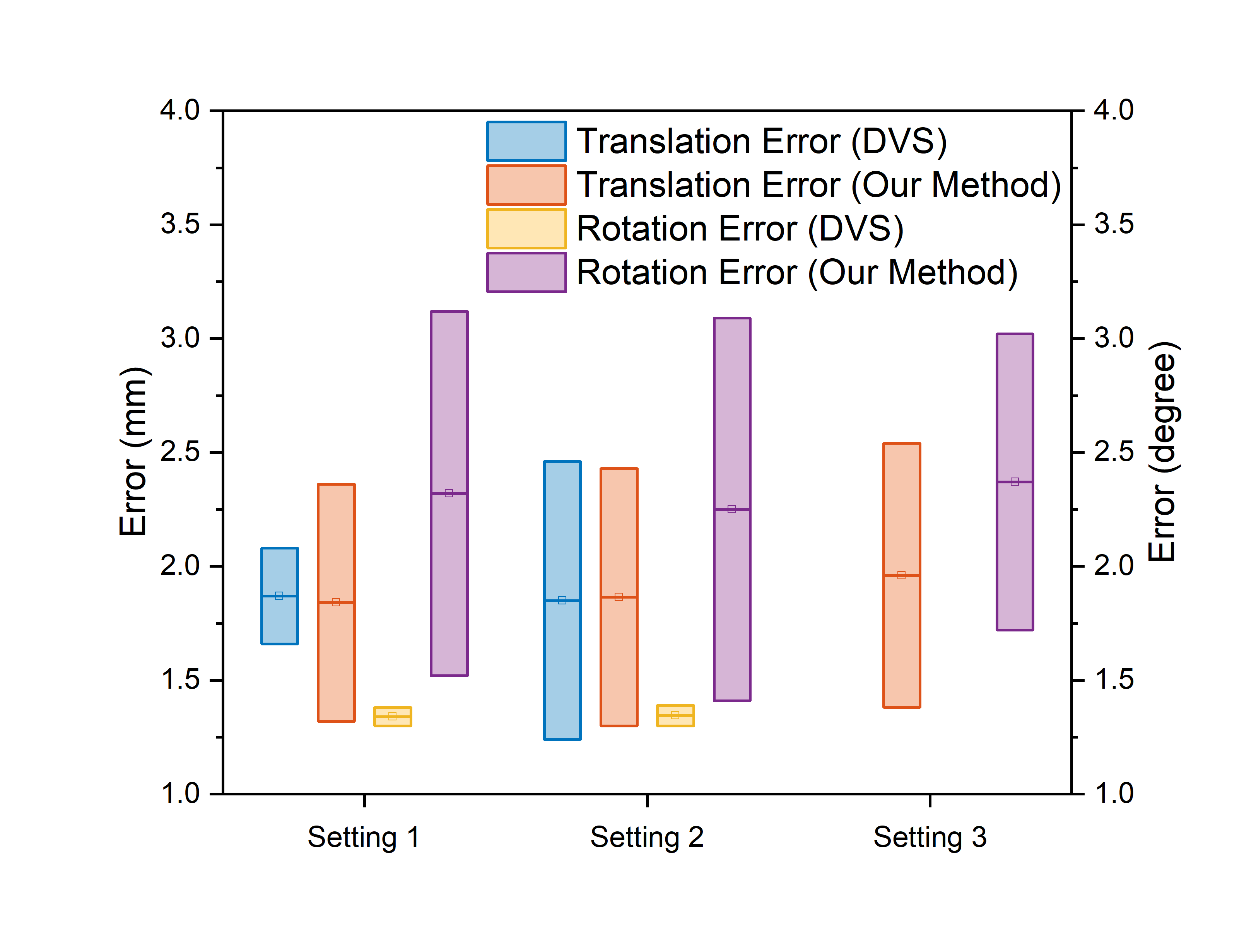}
		\caption{Translation and rotational error distributions of DVS and our method (error threshold = $2~\rm{mm}$) in different configuration settings.}
		\label{fig.comparison_our_and_dvs_error}
	\end{center}
\end{figure}

\begin{figure}[htbp]
	\begin{center}
    \includegraphics[width=8cm]{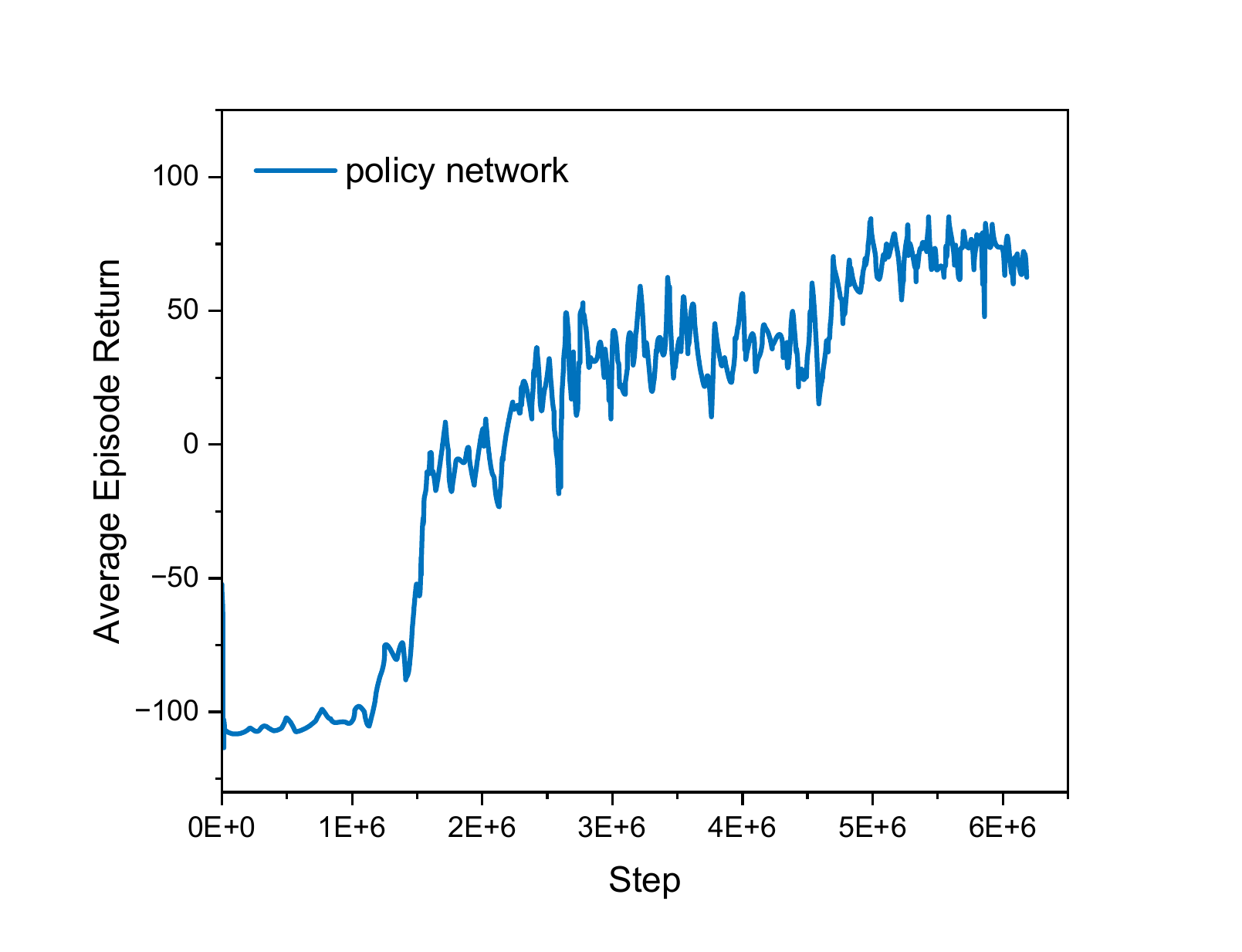}
		\caption{Training curve with average return per episode of proposed method in third experimental configuration.}
		\label{fig.vs-control-policy-training-reward-curve}
	\end{center}
\end{figure}
We also quantitatively evaluated with the error distributions in translation  and rotation of multi-perspective visual servoing in three experimental settings, as depicted in Fig.~\ref{fig.comparison_our_and_dvs_error}. The cases of failed convergence to the target state were shown as empty bars. The error threshold of $2~\rm{mm}$ was utilized during training our model. %
In simple scenarios of setting 1 and setting 2, we observed a slight increase in errors in proposed RL-based algorithm compared to DVS method. Our approach involves a trade-off between exploration and exploitation. Specifically, in simple scenarios, where problems are relatively straightforward, traditional algorithms may tend to exploit known information, while our reinforcement learning-based algorithm may lean towards exploring new possibilities. This tendency might lead to a minor increase in errors in simple scenarios. However, in complex scenarios, this exploratory strategy could be more advantageous in finding better solutions. Our algorithm demonstrates greater adaptability. It can learn and adjust in different scenarios. In complex situations, where there is significant environmental variation and uncertainty, traditional algorithms might struggle to adapt. In contrast, our algorithm excels in handling complexity and performs better in such scenarios. In summary, although our method may experience a slight increase in errors in simple scenarios, it exhibits outstanding performance in complex scenarios. This suggests that our approach is more robust and adaptable, enabling it to function effectively in the intricate environments of the real world.

Finally, we trained proposed method using HER, learning from demonstration, additional exploration in third experimental setting with more complicated scenes. The training took 141 hours, as demonstrated in Fig.~\ref{fig.vs-control-policy-training-reward-curve}. 

\section{CONCLUSIONS AND FUTURE WORK}\label{CONCLUSIONS}
We proposed a novel closed-loop multi-perspective reinforcement learning-based visual servoing network. HER, learning from demonstration and additional exploration methods were utilized to alleviate the pain of convergence in sparse reward function and sparse success behaviors in action space. The robot agent with velocity control was developed in simulation to perform visual servoing with different complicated scenarios. The robot actions were estimated based on latent space representations learned from visual states. 

The comparison experiments proved that our variant TD3+HER+Exploration+Demonstration demonstrated ability of our method in multi-perspective visual servoing using reinforcement learning. From quantitative experiments of our method and traditional method DVS, our method outperformed the DVS with a mean success rate of $97.0\%$ in different experimental settings. Meanwhile, our method could converge in complicated scenes with desired error distributions in translation and rotation. The mean translation errors of our method achieved performances of the success cases of accurate traditional method DVS. The future work will extend the reinforcement learning network in robot arm and five-finger hand manipulation.

% 
% \balance

\bibliographystyle{IEEEtran}
\bibliography{main}

\end{document}